\documentclass[10pt,twocolumn,letterpaper]{article}
\usepackage[pagenumbers]{cvpr}
\usepackage{times}
\usepackage{latexsym}
\usepackage{graphicx}
\usepackage{tabularx}
\usepackage{xspace}
\usepackage{xcolor}
\usepackage[table,xcdraw]{xcolor}
\usepackage{tcolorbox}
\usepackage{booktabs}
\usepackage{multirow}
\usepackage{amsmath} 
\usepackage{tabularx}
\usepackage{array}
\usepackage{amssymb}

\definecolor{cvprblue}{rgb}{0.21,0.49,0.74}
\definecolor{my_green}{RGB}{46, 151, 78}
\definecolor{my_red}{RGB}{216, 37, 34}
\usepackage[pagebackref,breaklinks,colorlinks,allcolors=cvprblue]{hyperref}
\usepackage{makecell}
\title{VideoArgus: Agentic Rubric-Grounded Unified Evaluation \\for Video Generation and Editing}

\author{Ziyun Zeng, Zixuan Wang, Yongsheng Yu, Hang Hua, Jiebo Luo\\
University of Rochester\\
{\tt\small \{zzeng24,zwang234,yyu90\}@ur.rochester.edu,\{hhua2,jluo\}@cs.rochester.edu}
}

\begin{document}
\maketitle
\begin{abstract}
Evaluating generated videos remains challenging because existing benchmarks rely on fixed evaluation content, cover only a subset of generation and editing settings, and provide limited evidence for their scores. We introduce \textbf{VideoArgus}, a unified rubric-grounded framework covering five video generation and editing settings. For each input instance, VideoArgus generates an output-blind, sample-specific rubric once and reuses it to evaluate all corresponding candidate videos. The rubric defines concrete criteria, scoring rules, failure modes, and evidence plans, which guide criterion-specific VLM QA and visual tools to produce evidence-grounded criterion scores, rationales, and a diagnostic report. We further construct \textbf{VideoArgus-Bench}, containing 1,026 curated input instances built from 653 high-quality images and 416 high-quality videos, with all benchmark rubrics pre-generated, frozen, and released. On a separate 1,260-video human-alignment set, VideoArgus achieves higher within-input Spearman and Kendall correlations with human judgments than the corresponding benchmark-specific evaluators across all five tasks. Model rankings also remain largely consistent across different rubric-generation and evaluation-VLM backbones. All code and data are released. Visit our  \href{https://zzzmyyzeng.github.io/VideoArgus}{project page}.
\end{abstract}
\begin{figure*}[t]
    \centering
    \includegraphics[width=\linewidth]{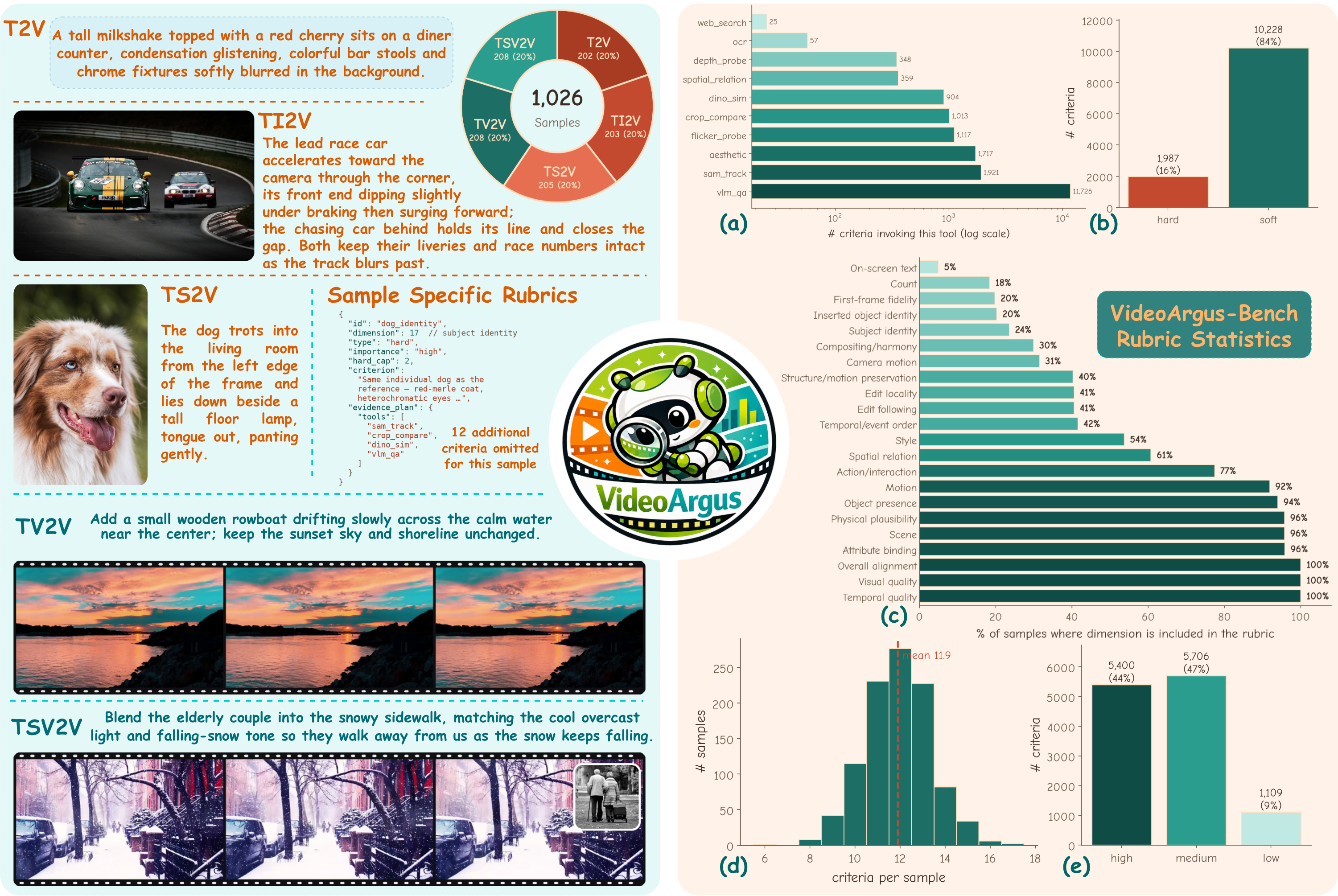}
    \caption{Overview of VideoArgus-Bench. Left: representative input instances from the five supported tasks---T2V, TI2V, TS2V, TV2V, and TSV2V---with an abridged visualization of the sample-specific rubric generated for one input instance. The inset summarizes the composition of the 1,026 benchmark instances. Right: rubric statistics over all input instances in VideoArgus-Bench: (a) the total number of criteria invoking each evaluation tool, shown on a logarithmic scale; (b) the counts and proportions of hard and soft criteria; (c) the percentage of input instances whose rubrics include each semantic dimension; (d) the distribution of the number of criteria per instance, with a mean of 11.9; and (e) the counts and proportions of high-, medium-, and low-importance criteria.}
    \label{fig:teaser}
\end{figure*}    
\section{Introduction}
\definecolor{okgreen}{HTML}{1B7F3B}   
\definecolor{nored}{HTML}{C0392B}     
\definecolor{pamber}{HTML}{B7791F}    

\newcommand{\cmark}{\textcolor{okgreen}{\textbf{\checkmark}}}
\newcommand{\xmark}{\textcolor{nored}{\ensuremath{\boldsymbol\times}}}
\newcommand{\pmark}{\textcolor{pamber}{\ensuremath{\boldsymbol\sim}}}

\newcommand{\vhead}[2]{%
\begin{tabular}[b]{@{}c@{}}
\textbf{#1}\\
\textbf{#2}
\end{tabular}}

\begin{table*}[t]
\centering
\small
\setlength{\tabcolsep}{2.2pt}
\renewcommand{\arraystretch}{1.10}
\resizebox{\textwidth}{!}{%
\begin{tabular}{@{}l|l|r|l|l|ccccc@{}}
\toprule
\textbf{Benchmark}
& \textbf{Task}
& \textbf{\#Items}
& \textbf{Eval. Content}
& \textbf{Scoring}
& \textbf{S-Spec.}
& \textbf{Adapt.}
& \textbf{VLM}
& \textbf{CV}
& \textbf{Interp.} \\
\midrule

VBench~\cite{huang2024vbench}
& T2V
& 1{,}746
& Fixed (1 of 16)
& Dimension
& \xmark & \xmark & \xmark & \cmark & \xmark \\

VBench-2.0~\cite{zheng2025vbench}
& T2V
& 1{,}155
& Fixed (1 of 18)
& Dimension
& \pmark & \xmark & \cmark & \cmark & \xmark \\

EvalCrafter~\cite{liu2024evalcrafter}
& T2V
& 700
& Fixed (all 17)
& Metric
& \xmark & \xmark & \xmark & \cmark & \xmark \\

T2V-CompBench~\cite{sun2025t2v}
& T2V
& 1{,}400
& Fixed (1 of 7)
& Dimension
& \xmark & \xmark & \cmark & \cmark & \xmark \\

PhyGenBench \citep{meng2024towards}
& T2V
& 160
& Input-derived (var.)
& Question
& \cmark & \pmark & \cmark & \cmark & \pmark \\

VideoGen-Eval~\cite{yang2025videogen}
& T2V, I2V
& 700
& Input-derived (var.)
& Dimension
& \cmark & \pmark & \cmark & \cmark & \cmark \\

\midrule

VBench-I2V~\cite{huang2024vbench}
& TI2V
& 1{,}118
& Fixed (all 9)
& Dimension
& \xmark & \xmark & \xmark & \cmark & \xmark \\

OpenS2V-Nexus~\cite{yuan2026opens2v}
& TS2V
& 180
& Fixed (all 6)
& Dimension
& \xmark & \xmark & \cmark & \cmark & \xmark \\

\midrule

OpenVE-Bench~\cite{he2025openve}
& TV2V
& 431
& Fixed (all 8)
& Dimension
& \xmark & \xmark & \cmark & \xmark & \xmark \\

FiVE-Bench~\cite{li2025five}
& TV2V
& 420
& Input-derived (1)
& Question
& \cmark & \xmark & \cmark & \cmark & \pmark \\

IVEBench~\cite{chen2025ivebench}
& TV2V
& 600
& Fixed (var. of 12)
& Metric
& \xmark & \pmark & \cmark & \cmark & \xmark \\

EditVerseBench~\cite{ju2025editverse}
& TV2V, TSV2V
& 200
& Fixed (all 4)
& Dimension
& \xmark & \xmark & \cmark & \cmark & \xmark \\

\midrule

VACE-Bench~\cite{jiang2025vace}
& \shortstack[l]{TI2V, TS2V,\\TV2V, TSV2V}
& 480
& Fixed (all 8)
& Metric
& \xmark & \xmark & \xmark & \cmark & \xmark \\

\midrule
UniVBench$^{\star}$~\cite{wei2026univbench}
& \shortstack[l]{T2V, TS2V,\\TV2V, TSV2V}
& 620
& Fixed (all 21)
& Dimension
& \pmark & \pmark & \cmark & \pmark & \cmark \\

\midrule

\textbf{VideoArgus (ours)}
& \shortstack[l]{\textbf{T2V, TI2V, TS2V,}\\
                 \textbf{TV2V, TSV2V}}
& \textbf{1{,}026}
& \textbf{Input-derived (var.)}
& \textbf{Criterion}$^{\dagger}$
& \cmark & \cmark & \cmark & \cmark & \cmark \\

\bottomrule
\end{tabular}
}
\vspace{-2mm}
\caption{
Comparison with representative video generation and editing benchmarks.
\emph{Eval. Content} indicates whether evaluated properties are predefined (\emph{Fixed}) or instantiated from each input (\emph{Input-derived}); parentheses summarize their per-item use.
\emph{Scoring} denotes the finest independently scored unit.
\emph{S-Spec.} denotes sample-specific evaluation content. 
\emph{Adapt.} indicates whether evidence acquisition or tool execution is adapted to individual scoring units.
\emph{VLM}, \emph{CV}, and \emph{Interp.} denote VLM judging, specialized visual tools, and diagnostic feedback.
\cmark~= yes, \pmark~= partial, and \xmark~= no.
$^{\star}$Only tasks relevant to our comparison are reported.
$^{\dagger}$Criterion-level evaluation uses dimensions only as taxonomy labels and instantiates zero, one, or multiple independently scored criteria, each with its own requirement, scoring rule, and evidence plan. Dimension-level evaluation assigns a scoring slot to a semantic category. 
}
\vspace{-2mm}
\label{tab:video_eval_comparison}
\end{table*}

Modern video models support text-to-video generation (T2V), text-and-image-driven video generation (TI2V), text-and-subject-driven video generation (TS2V), text-driven video editing (TV2V), and text-and-subject-driven video editing (TSV2V). Therefore, a unified protocol is needed to evaluate these heterogeneous generation and editing settings consistently.

Existing evaluators remain fragmented across tasks and often use predefined dimensions shared by all inputs. Yet requirements such as counting, visible text, temporal order, subject identity, and source preservation are instance-dependent. Specialized metrics measure narrow properties, while holistic VLM judges often combine multiple requirements with limited supporting evidence, making failures difficult to localize.

We introduce \textbf{VideoArgus}, a unified rubric-grounded framework for the five generation and editing settings above. For each input instance, VideoArgus generates an output-blind, sample-specific rubric once and reuses it to evaluate all corresponding candidate videos. The rubric decomposes the input requirements into independently scored criteria with explicit scoring rules and evidence plans.
For each candidate video, VideoArgus executes the evidence plan associated with each criterion, selectively invoking criterion-specific VLM QA and visual tools. The resulting evidence is used to produce criterion-level scores and rationales before aggregation into a final score and diagnostic report.

We further instantiate this framework at scale through \textbf{VideoArgus-Bench}, a unified benchmark covering all five tasks. The benchmark contains 1,026 input instances constructed from 653 unique images and 416 unique videos, covering T2V, TI2V, TS2V, TV2V, and TSV2V. Input instances are automatically authored from task-specific blueprints and visual inputs, then verified and revised by human reviewers. All sample-specific rubrics are pre-generated and frozen for reproducible model comparison. The rubric-generation procedure is not restricted to VideoArgus-Bench: users may apply it to new input instances and reuse the resulting rubric to evaluate any number of candidate videos for the corresponding input.

We evaluate 55 model--task pairs across the five generation and editing settings on VideoArgus-Bench. We further evaluate VideoArgus on a separate human-alignment set of 1,260 videos scored by 15 annotators. Across all five tasks, VideoArgus achieves higher within-input Spearman and Kendall correlations with human judgments than the corresponding benchmark-specific evaluators. Tool ablations show that specialized visual tools provide complementary evidence beyond criterion-specific VLM inspection. Moreover, model rankings remain largely consistent across different rubric-generation and evaluation-VLM backbones.
Our contributions are summarized as follows:
\begin{itemize}
    \item We introduce an output-blind rubric-generation mechanism that constructs a reusable sample-specific evaluation specification once per input instance. Each rubric defines independently scored criteria with explicit scoring rules, failure modes, and evidence plans, and is reused across all candidate outputs for that input. The same generation procedure can be applied to new user-provided input instances beyond VideoArgus-Bench.

    \item We propose a rubric-grounded video evaluation procedure that selectively executes criterion-specific visual tools and produces evidence-grounded scores, rationales, and diagnostic reports. The rubric-generation and evaluation-VLM backbones can be replaced independently.
    
    \item We construct VideoArgus-Bench, with 1,026 input instances across five generation and editing tasks, evaluate 55 model--task pairs, and demonstrate stronger human alignment and robustness to different rubric-generation and evaluation-VLM backbones.

\end{itemize}
\section{Related Work}
\label{sec:related_work}

\paragraph{Video Generation and Editing.}
Recent video foundation models have substantially advanced text- and image-conditioned synthesis through large-scale diffusion transformers, improved training data, and model scaling~\cite{yang2025cogvideox,wu2025hunyuanvideo,wan2025wan,ma2025step,li2026skyreels}. Reference-conditioned methods further enable consistent generation from single or multiple subject images~\cite{deng2025magref,li2025bindweave,zhang2025kaleido,wang2026refalign,yuan2026opens2v}. In parallel, video editing has expanded from text-guided manipulation to reference-guided editing, while emerging unified architectures increasingly support both generation and editing within a single model~\cite{hacohen2024ltx,wei2025univideo,yu2026aurora,yang2026omni,pan2026omniweaving}. As these capabilities converge, a unified protocol is needed to evaluate heterogeneous generation and editing settings consistently.

\paragraph{Video Generation and Editing Benchmarks.}
Existing video benchmarks are largely organized around individual task families. Early T2V benchmarks such as VBench~\cite{huang2024vbench}, EvalCrafter~\cite{liu2024evalcrafter}, and T2V-CompBench~\cite{sun2025t2v} assess semantic alignment, compositional correctness, motion, and perceptual quality, while PhyGenBench~\cite{meng2024towards} and VBench-2.0~\cite{zheng2025vbench} emphasize physical plausibility and intrinsic faithfulness. Task-specific extensions cover additional conditioning and editing scenarios: VBench-I2V~\cite{huang2024vbench} targets first-frame-conditioned generation, OpenS2V-Nexus~\cite{yuan2026opens2v} evaluates reference-subject consistency, and OpenVE-Bench~\cite{he2025openve} and EditVerseBench~\cite{ju2025editverse} focus on instruction following and source preservation. Several recent efforts evaluate multiple generation and editing settings within a shared framework~\cite{wei2026univbench,yang2025videogen}. VideoArgus-Bench extends this direction by unifying five representative settings in a single curated benchmark and associating every instance with a pre-generated, frozen, sample-specific rubric.

\paragraph{VLM-based, Adaptive, and Interpretable Evaluation.}
Automatic video evaluation broadly follows three lines of development. Specialized metrics provide reproducible but narrow measurements of properties such as text--video alignment, perceptual quality, motion, temporal consistency, tracking, and reference similarity~\cite{huang2024vbench,liu2024evalcrafter,ju2025editverse,zeng2025use}. VLM-based evaluators instead reason jointly over instructions, conditioning inputs, and sampled video frames, enabling more holistic assessment of compositional correctness, physical plausibility, subject fidelity, and edit following~\cite{sun2025t2v,meng2024towards,yuan2026opens2v,he2025openve,zeng2026automated}. More recently, multimodal approaches have advanced generation and editing, and agentic reasoning, memory, and tool use~\cite{hua2024mmcomposition,hua2025mmigbench,wang2024dancecamanimator,wang2024dancecamera3d,wang2022groupdancer,yu2026aurora,yu2025omnipaint,zeng2026mementogui,zeng2026mira,hu2022promptcap,lin2023videoxum,tang2025video}, while recent evaluators adapt questions, criteria, and evidence procedures
to individual inputs~\cite{zheng2025vbench,wei2026univbench,yang2025videogen}. 
\section{VideoArgus}
\begin{figure*}
    \centering
    \includegraphics[width=0.98\linewidth]{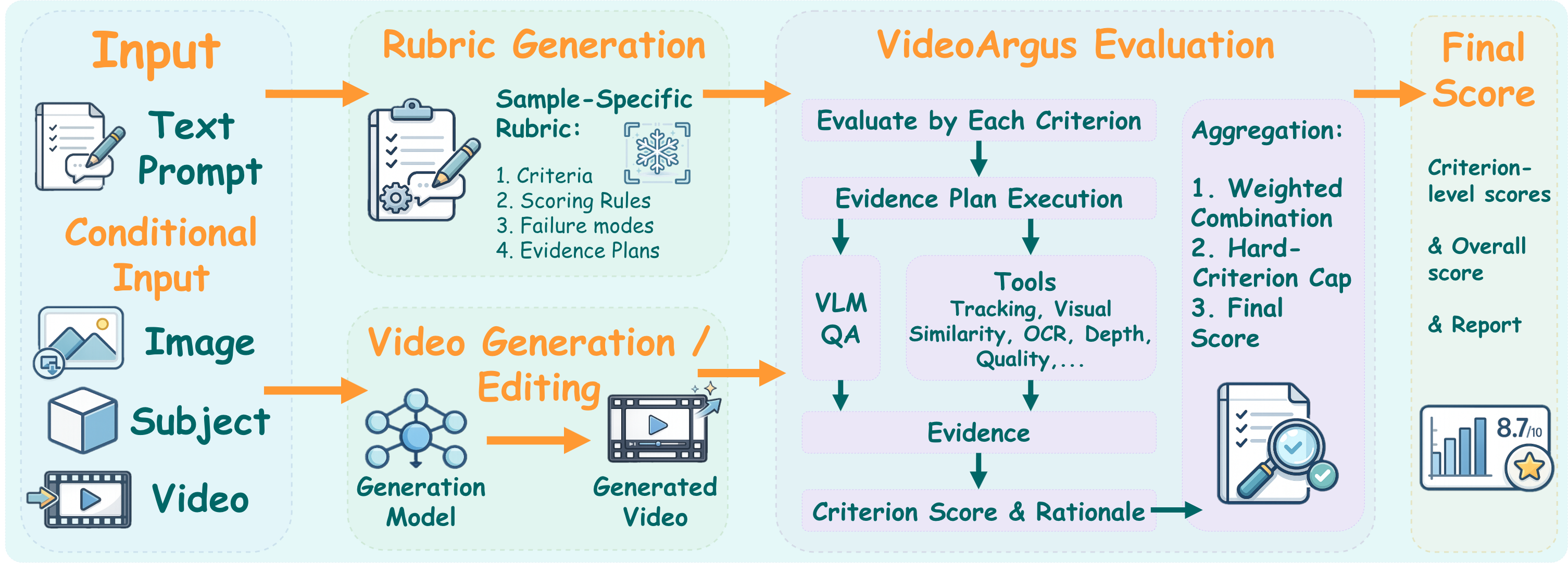}
    \caption{Overview of VideoArgus. An input instance consists of a text prompt and optional image, subject, or video conditions. The input-side rubric generator observes only this instance and constructs an output-blind, sample-specific rubric once, specifying concrete criteria, scoring rules, expected failure modes, and evidence plans. The resulting frozen rubric is reused to evaluate candidate videos generated or edited from the same input. For each criterion, VideoArgus executes its evidence plan using VLM QA and, when applicable, specialized tools such as tracking, visual similarity, OCR, depth analysis, and perceptual-quality assessment. The collected evidence supports a criterion-level score and rationale. Finally, criterion scores are importance-weighted and aggregated with hard-criterion caps to produce the overall score and diagnostic report.}
    \label{fig:videoargus_pipeline}
\end{figure*}
VideoArgus first generates an output-blind, sample-specific rubric once for each input instance and then reuses the frozen rubric to evaluate all candidate videos associated with that input. As illustrated in Fig.~\ref{fig:videoargus_pipeline}, rubric generation depends only on the input instance, whereas video evaluation combines the fixed rubric with each candidate output. This design keeps the evaluation specification identical across competing outputs, amortizes rubric generation as additional models are evaluated, and allows the rubric-generation and evaluation-VLM backbones to be replaced independently.

\subsection{Reusable Sample-Specific Rubric Generation}
\begin{table*}[t]
\centering
\setlength{\tabcolsep}{2.4pt}
\renewcommand{\arraystretch}{1.06}

\resizebox{\textwidth}{!}{%
\begin{tabular}{@{}
l l r @{\hspace{1.0em}}
l l r @{\hspace{1.0em}}
l l r @{\hspace{1.0em}}
l l r @{\hspace{1.0em}}
l l r
@{}}
\toprule

\multicolumn{3}{c}{\textbf{T2V}} &
\multicolumn{3}{c}{\textbf{TI2V}} &
\multicolumn{3}{c}{\textbf{TS2V}} &
\multicolumn{3}{c}{\textbf{TV2V}} &
\multicolumn{3}{c}{\textbf{TSV2V}} \\

\cmidrule(r){1-3}
\cmidrule(lr){4-6}
\cmidrule(lr){7-9}
\cmidrule(lr){10-12}
\cmidrule(l){13-15}

\textbf{Model} & \textbf{\#P} & \textbf{Score} &
\textbf{Model} & \textbf{\#P} & \textbf{Score} &
\textbf{Model} & \textbf{\#P} & \textbf{Score} &
\textbf{Model} & \textbf{\#P} & \textbf{Score} &
\textbf{Model} & \textbf{\#P} & \textbf{Score} \\

\midrule

Seedance 2.0 & -- & \textbf{9.34} &
Seedance 2.0 & -- & \textbf{8.97} &
Seedance 2.0 & -- & \textbf{9.16} &
Seedance 2.0 & -- & \textbf{7.54} &
Seedance 2.0 & -- & \textbf{8.10} \\

Veo 3.1 & -- & \underline{8.66} &
Kling v3 Omni & -- & \underline{8.41} &
Kling v3 Omni & -- & \underline{8.22} &
Kling v3 Omni & -- & \underline{6.79} &
Kling v3 Omni & -- & \underline{7.16} \\

Kling v3 Omni & -- & 8.61 &
Veo 3.1 & -- & 8.36 &
OmniWeaving & 8.3B & 7.28 &
Kiwi-Edit & 5B & 6.18 &
Aurora & 5B & 5.03 \\

HunyuanVideo-1.5 & 8.3B & 8.54 &
HunyuanVideo-1.5 & 8.3B & 8.11 &
SkyReels-V3-R2V & 14B & 7.25 &
Aurora & 5B & 6.02 &
OmniWeaving & 8.3B & 4.97 \\

Wan2.2-T2V & 14B & 8.41 &
OmniWeaving & 8.3B & 7.99 &
Veo 3.1 & -- & 6.97 &
OmniWeaving & 8.3B & 5.87 &
Kiwi-Edit & 5B & 4.51 \\

OmniWeaving & 8.3B & 8.40 &
Wan2.2-TI2V & 5B & 7.82 &
Phantom & 14B & 6.89 &
ReCo & 1.3B & 5.77 &
VACE-Fun & 14B & 4.20 \\

LongCat-Video & 13.6B & 8.11 &
CogVideoX & 5B & 7.29 &
VACE & 14B & 6.82 &
VideoCoF & 14B & 5.53 &
UniVideo & 13B & 3.20 \\

Step-Video & 29B & 7.77 &
CogVideoX1.5 & 5B & 7.16 &
HunyuanCustom & 13B & 6.75 &
LucyEdit & 5B & 4.99 &
VACE & 1.3B & 2.81 \\

SANA-Video & 2B & 7.54 &
Step-Video & 29B & 7.11 &
Wan2.2-TI2V & 5B & 5.67 &
VACE-Fun & 14B & 4.92 &
AnyV2V & 1.4B & 2.66 \\

CogVideoX1.5 & 5B & 7.52 &
LTX-Video & 2B & 6.25 &
UniVideo & 13B & 4.13 &
LTX-Video & 2B & 3.48 &
VACE & 14B & 2.62 \\

CogVideoX & 5B & 7.39 &
SANA-Video & 2B & 5.89 &
LTX-Video & 2B & 3.89 &
UniVideo & 13B & 3.36 &
LTX-Video & 2B & 2.45 \\

\bottomrule
\end{tabular}%
}
\vspace{-2mm}
\caption{
VideoArgus leaderboards for video generation and editing tasks. Each task ranks the evaluated generators by their VideoArgus score. The best result per task is shown in \textbf{bold}, and the second-best result is \underline{underlined}. \#P denotes model parameter size.
}
\vspace{-2mm}
\label{tab:videoargus_leaderboard}
\end{table*}
The requirements of a generated video depend on its text prompt and visual conditioning inputs. Counting, visible text, subject identity, and source preservation may be critical for one instance but irrelevant to another. We therefore represent each input instance as
\begin{equation}
x_i = \left(p_i,I_i^{\mathrm{first}},\mathcal{I}_i^{\mathrm{subj}},V_i^{\mathrm{src}}\right),
\end{equation}
where $p_i$ is the textual instruction and the remaining entries denote an optional first-frame image, subject images, and source video. This representation supports T2V, TI2V, TS2V, TV2V, and TSV2V under a single interface.

The textual instruction is passed without rewriting. Images preserve their aspect ratio, with the longest side limited to 768 pixels. Source videos are sampled at 2 fps. A single rubric-generation prompt is used across all tasks, while applicable criteria are determined from the conditioning inputs present in each instance.

Crucially, rubric generation only observes the input instance:
\begin{equation}
\mathcal{R}_i = G_{\mathrm{rubric}}(x_i).
\end{equation}
Each rubric $R_i$ is generated once, indexed only by the input instance, and reused for every candidate output $\{\hat{V}_{i,m}\}_{m=1}^{M}$. This output-blind design prevents the evaluation specification from adapting to the strengths or failures of a particular model and ensures that all outputs for the same input are judged against identical requirements. For VideoArgus-Bench, these rubrics are pre-generated, frozen, and released, so benchmark users execute only the video-evaluation stage. For a new user-provided input, the same procedure generates a new rubric once, which can then be reused across all candidate videos for that input.

The rubric generator considers a unified pool of 22 general and input-conditional dimensions. These dimensions cover object and attribute correctness, counting, spatial and temporal relations, actions, motion, camera behavior, physical plausibility, visual quality, style, first-frame fidelity, subject identity, edit following, source preservation, and compositing quality. The dimension pool serves as a semantic taxonomy rather than a one-slot-per-dimension template. For each input instance, a dimension may instantiate zero, one, or multiple independently scored criteria, depending on the distinct observable requirements expressed by the input.

Each criterion specifies a semantic dimension, a concrete requirement, a hard or soft constraint type, an importance level, an integer 0--10 scoring rule, expected failure modes, an optional hard cap, and an evidence plan. The evidence plan identifies the target entity or event, the available reference source, and an ordered sequence of tools for collecting relevant evidence. This structured representation turns the rubric into an executable evaluation specification rather than a generic question list. Separately scoring distinct observable requirements prevents multiple requirements from being coupled within a single judgment and provides criterion-specific evidence and diagnostic feedback.

\subsection{Rubric-Grounded Video Evaluation}
Given a candidate video $\hat{V}$ and the frozen sample-specific rubric $R=\{r_i\}_{i=1}^{K}$ for its input instance, VideoArgus evaluates each criterion $r_i$ independently using its associated evidence plan $\pi_i$. The plan identifies the relevant reference inputs, target entities or events, and the sequence of tools used to evaluate the criterion. Criterion-specific execution allows heterogeneous requirements to be assessed using different forms of visual evidence within the same evaluation protocol.

All evidence operations share a unified tool interface. \texttt{vlm\_qa} serves as a general-purpose tool for semantic questions rather than as a separate holistic evaluator. Specialized tools support object localization and tracking, reference-guided cropping, DINOv3-based visual similarity, OCR, depth and spatial-relation analysis, temporal flicker detection, perceptual-quality assessment, and optional visual-reference retrieval. Each criterion invokes only the tools specified by its evidence plan.

To support reliable execution across heterogeneous input instances, VideoArgus applies a plan-normalization layer before tool execution. It resolves visual conditions, checks tool arguments and inter-tool dependencies, and maps unavailable operations to supported alternatives. When specialized evidence cannot be obtained, the evaluator falls back to a criterion-specific \texttt{vlm\_qa} query. The executed tool sequence and any fallback decisions are retained in the evaluation report.

After executing the evidence plan $\pi_i$, VideoArgus obtains structured evidence $e_i$. The criterion judge receives the criterion specification $r_i$, the collected evidence $e_i$, and representative frames from $\hat{V}$, and returns
\begin{equation}
J(r_i,e_i,\hat{V})=(s_i,q_i),
\end{equation}
where $s_i\in\{0,\ldots,10\}$ is the criterion score and $q_i$ is an evidence-grounded rationale. Criterion importance, failure thresholds, and hard caps are hidden from the judge and used only during aggregation. If no valid score can be parsed, the criterion is excluded from the aggregation.
\begin{figure}[t]
    \centering
    \includegraphics[width=\linewidth]{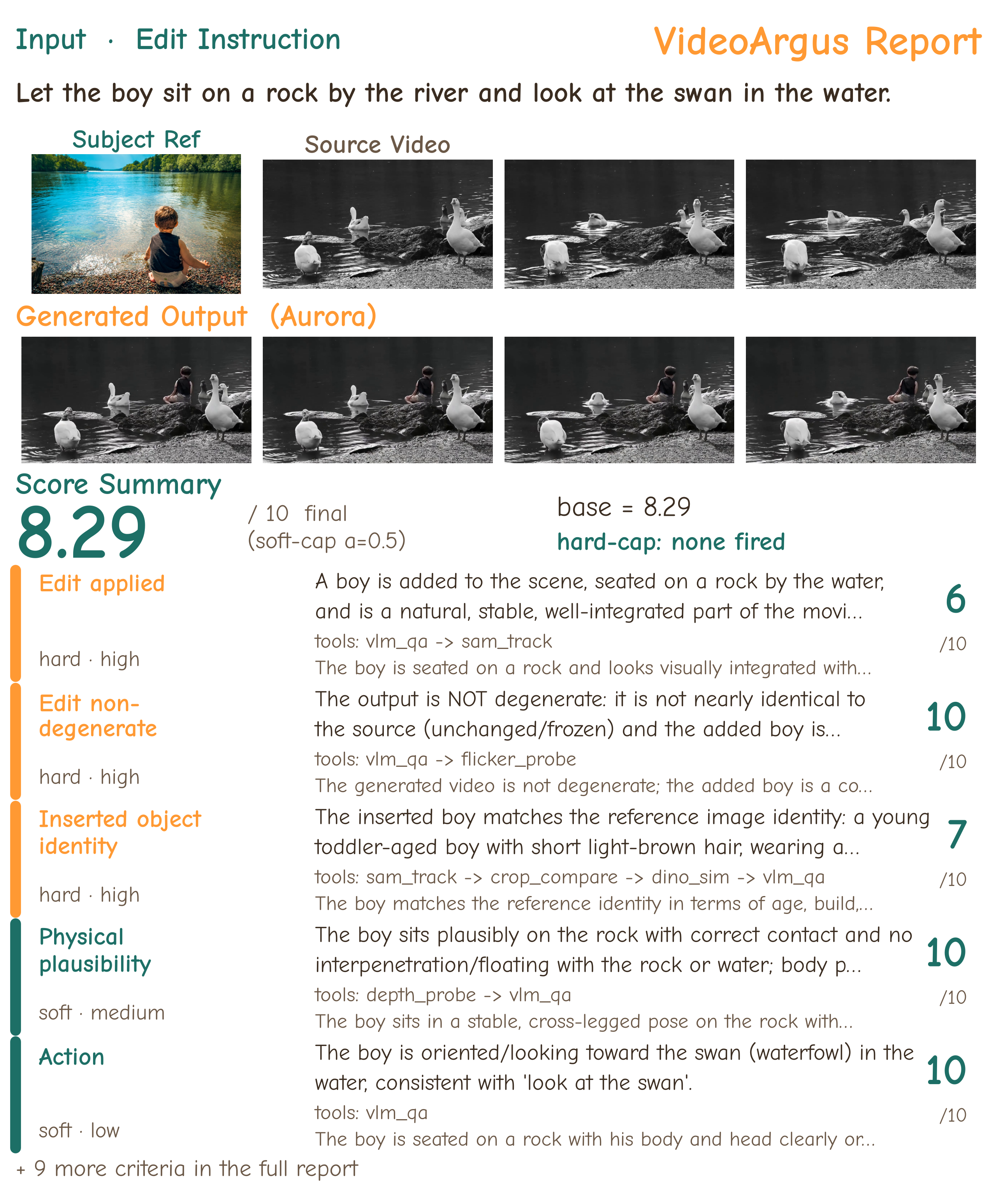}
    \caption{Example VideoArgus report for a TSV2V output by Aurora~\cite{yu2026aurora}, showing the input, generated frames, aggregate score, and five selected criterion results with tool traces, evidence-grounded rationales, and scores. The full report contains all fourteen criteria and aggregation metadata.}
\end{figure}
Let $\mathcal{I}(\hat{V})$ denote the indices of criteria with valid scores, and let $w_i\in\{1,2,3\}$ correspond to low, medium, and high importance, respectively. The importance-weighted base score is
\begin{equation}
B(\hat{V})=\frac{\sum_{i\in\mathcal{I}(\hat{V})} w_i s_i}{\sum_{i\in\mathcal{I}(\hat{V})} w_i}.
\end{equation}
During rubric generation, each hard criterion is assigned an individual failure threshold $t_i$ and a criterion-specific score cap $c_i$. Its cap is activated when the criterion score falls below its threshold. The set of activated hard criteria is
\begin{equation}
\mathcal{F}(\hat{V})=\left\{i\in\mathcal{I}(\hat{V})\;\middle|\;r_i \text{ is hard and } s_i<t_i\right\}.
\end{equation}
When multiple hard criteria are activated, the most restrictive cap is used:
\begin{equation}
\kappa(\hat{V})=\min\left(\{c_i\mid i\in\mathcal{F}(\hat{V})\}\cup\{10\}\right).
\end{equation}
The final score is
\begin{equation}
S(\hat{V})=(1-\alpha)B(\hat{V})+\alpha\min\!\left(B(\hat{V}),\kappa(\hat{V})\right),
\end{equation}
where $\alpha=0.5$. This aggregation penalizes essential failures while retaining information from all criterion scores.

Rubric generation and rubric-grounded evaluation use separate model interfaces. The rubric generator observes only the input instance and is executed once per input. For each candidate video, the evaluation VLM performs criterion-specific \texttt{vlm\_qa} and final criterion judgment using the frozen rubric and collected tool evidence. The two model backbones can therefore be selected or replaced independently.

\section{VideoArgus-Bench}
We construct VideoArgus-Bench, a unified benchmark spanning T2V, TI2V, TS2V, TV2V, and TSV2V, to instantiate VideoArgus at scale. Input instances are created from task-specific blueprints and, when required, quality-filtered visual assets collected from Pixabay~\cite{pixabay}. Candidate instructions are grounded in the actual conditioning inputs and automatically checked for feasibility, text--visual consistency, and duplication. Human reviewers then verify the resulting instances, revise unclear or impractical instructions, and discard invalid cases. Sample-specific rubrics are generated once from the finalized input instances using the output-blind procedure described above.

VideoArgus-Bench contains 1,026 input instances constructed from 653 unique images and 416 unique videos. Instructions contain approximately 26 words on average. The released rubrics are pre-generated and frozen, ensuring that all candidate outputs for the same input are evaluated against an identical specification. They contain 11.9 criteria per instance on average, ranging from 6 to 17. Figure~\ref{fig:teaser} summarizes the task composition and rubric statistics.
\section{Experiments}
We benchmark representative models on VideoArgus-Bench, evaluate agreement with human judgments, and analyze specialized-tool contributions and backbone consistency.

\subsection{Experimental Setup}
\paragraph{Evaluated Models.}
We evaluate the API-based models, including Veo 3.1~\cite{googledeepmind2026veo3modelcard}, Seedance 2.0~\cite{seedance2026seedance}, and Kling v3 Omni~\cite{kuaishou2026kling30omni}. We also evaluate a broad set of open-source models, including Wan 2.2~\cite{wan2025wan}, HunyuanVideo-1.5~\cite{wu2025hunyuanvideo}, Step-Video~\cite{ma2025step,huang2025step}, LongCat-Video~\cite{team2025longcat}, SkyReels-V3~\cite{li2026skyreels}, OmniWeaving~\cite{pan2026omniweaving}, SANA-Video~\cite{chen2025sana}, CogVideoX~\cite{yang2025cogvideox}, CogVideoX1.5~\cite{yang2025cogvideox}, LTX-Video~\cite{hacohen2024ltx}, Phantom~\cite{liu2025phantom}, VACE~\cite{wan2025wan}, HunyuanCustom~\cite{hu2025hunyuancustom}, UniVideo~\cite{wei2025univideo}, Aurora~\cite{yu2026aurora}, Kiwi-Edit~\cite{lin2026kiwi}, ReCo~\cite{zhang2025region}, VideoCoF~\cite{yang2026videocof}, LucyEdit~\cite{decart2025lucyedit}, VACE-Fun~\cite{alibabapai2025wan22vacefun}, AnyV2V~\cite{ku2024anyv2v}.
\paragraph{Benchmark Evaluation.}
For each input instance, all candidate videos share the same frozen rubric generated with Claude Opus~4.8. During rubric-grounded evaluation, Qwen3.6-27B, served with vLLM, is used for both criterion-specific \texttt{vlm\_qa} evidence collection and final criterion judgment. Specialized visual tools provide additional evidence when invoked by the rubric. Candidate videos are sampled at 4~fps with at most 24 frames. Dense temporal probes use up to 48 frames at 8 fps, and up to 12 representative frames are provided for final criterion judgment.

\paragraph{Human-alignment Set.}
We construct a separate set of 210 input instances: 48 T2V instances from VBench~\cite{huang2024vbench}, 48 TI2V instances from VBench-I2V~\cite{huang2024vbench}, 48 TS2V instances from OpenS2V-Nexus~\cite{yuan2026opens2v}, 56 TV2V instances from OpenVE-Bench~\cite{he2025openve} and 10 TSV2V instances from EditVerseBench~\cite{ju2025editverse}. Six model outputs per input yield 1,260 videos. Because the inputs come from external benchmarks, this set evaluates VideoArgus beyond VideoArgus-Bench. Fifteen trained annotators viewed the prompt and visual conditioning inputs, scored anonymized and randomized videos from 0 to 10, ranked ties, and their scores were averaged. For each task, we use the evaluator prescribed by the source benchmark with its released metrics, prompts, judge model, and aggregation.

\paragraph{Agreement metrics.}
\begin{table}[t]
\centering
\setlength{\tabcolsep}{1.5mm}

\resizebox{\columnwidth}{!}{%
\begin{tabular}{l|l|ccc}
\toprule
\textbf{Task} & \textbf{Method}
& $\text{R-}\rho$ & $\text{R-}\tau$ & $\text{P-}\rho$ \\
\midrule

\multirow[c]{3}{*}{T2V}
& Source Evaluation
& 0.162 & 0.104 & 0.073 \\
& VideoArgus-VLM
& \underline{0.436} & \underline{0.472} & \textbf{0.513} \\
& VideoArgus-Full
& \textbf{0.496} & \textbf{0.550} & \underline{0.480} \\
\midrule

\multirow[c]{3}{*}{TI2V}
& Source Evaluation
& -0.060 & -0.059 & -0.050 \\
& VideoArgus-VLM
& \underline{0.590} & \underline{0.673} & \underline{0.606} \\
& VideoArgus-Full
& \textbf{0.605} & \textbf{0.690} & \textbf{0.663} \\
\midrule

\multirow[c]{3}{*}{TS2V}
& Source Evaluation
& 0.524 & 0.426 & 0.471 \\
& VideoArgus-VLM
& \underline{0.612} & \textbf{0.692} & \underline{0.611} \\
& VideoArgus-Full
& \textbf{0.631} & \underline{0.670} & \textbf{0.632} \\
\midrule

\multirow[c]{3}{*}{TV2V}
& Source Evaluation
& 0.548 & 0.452 & 0.521 \\
& VideoArgus-VLM
& \underline{0.593} & \underline{0.601} & \textbf{0.636} \\
& VideoArgus-Full
& \textbf{0.608} & \textbf{0.625} & \underline{0.618} \\
\midrule

\multirow[c]{3}{*}{TSV2V}
& Source Evaluation
& 0.686 & 0.587 & \textbf{0.729} \\
& VideoArgus-VLM
& \underline{0.715} & \underline{0.751} & 0.703 \\
& VideoArgus-Full
& \textbf{0.749} & \textbf{0.829} & \underline{0.708} \\
\bottomrule
\end{tabular}%
}
\vspace{-2mm}
\caption{
Correlation with human judgment on five video generation and editing tasks. For each task, the corresponding benchmark-specific evaluator, VideoArgus-VLM, and VideoArgus-Full are separately correlated with judgments from the same 15 annotators on an identical set of videos. We report ranking Spearman ($\text{R-}\rho$), Kendall $\tau$ ($\text{R-}\tau$), and pooled Spearman ($\text{P-}\rho$). The best result per task is \textbf{bold}, and the second best is \underline{underlined}.
}
\vspace{-2mm}
\label{tab:human-correlation}
\end{table}

Our primary metric is within-instance ranking Spearman correlation ($R$-$\rho$). For each input instance, we compute the correlation between human and automatic scores across the six candidate models and then macro-average across instances
$R\text{-}\rho=\frac{1}{N}\sum_{i=1}^{N}\rho\left(\mathbf{s}_{i}^{\mathrm{human}},\mathbf{s}_{i}^{\mathrm{eval}}\right).$
This protocol measures whether an evaluator recovers the ordering of candidate models for the same input, without conflating differences in difficulty or score scale across input instances. We also report within-instance Kendall correlation ($R$-$\tau$) and pooled Spearman correlation ($P$-$\rho$). Instances with undefined correlations due to constant scores are excluded.

\subsection{Leaderboard on VideoArgus-Bench}

Table~\ref{tab:videoargus_leaderboard} report the mean VideoArgus score for each evaluated model and task. Within each input instance, all candidate outputs are evaluated using the same frozen rubric and evaluator configuration.  Seedance~2.0 achieves the highest score across all five tasks. Veo~3.1 ranks second on T2V, while Kling v3 Omni ranks second on TI2V, TS2V, TV2V, and TSV2V. Among open-source models, HunyuanVideo-1.5 performs strongest on T2V and TI2V, OmniWeaving leads TS2V, Kiwi-Edit leads TV2V, and Aurora leads TSV2V. The relative performance of open-source models varies across generation and editing settings, reflecting differences in requirements such as first-frame fidelity, subject preservation, edit following, and source consistency. Hence, a unified benchmark enables comparison across tasks while retaining the input-specific requirements of each setting.

\subsection{Agreement with Human Judgment}
\begin{table}[t]
\centering
{\small
\setlength{\tabcolsep}{1.1mm}
\renewcommand{\arraystretch}{1.15}
\begin{tabular}{l|c|cc|c|cc}
\toprule
\multirow{2}{*}{\textbf{Task}} & \multirow{2}{*}{\textbf{\#In.}}
& \multicolumn{2}{c|}{\textbf{Criteria / input}}
& \multirow{2}{*}{\textbf{Ratio}}
& \multicolumn{2}{c}{\textbf{Hard / input}} \\
 & & \textbf{Human} & \textbf{Induced} & & \textbf{Human} & \textbf{Induced} \\
\midrule
T2V   & 48 & 4.60 & 10.60 & 2.30$\times$ & 2.31 & 2.15 \\
TI2V  & 48 & 5.06 & 11.02 & 2.18$\times$ & 2.40 & 1.67 \\
TS2V  & 48 & 5.27 & 12.46 & 2.36$\times$ & 2.81 & 1.98 \\
TV2V  & 56 & 4.96 & 11.52 & 2.32$\times$ & 1.91 & 2.18 \\
TSV2V & 10 & 5.00 & 12.80 & 2.56$\times$ & 1.90 & 2.50 \\
\midrule
\textbf{All} & 210 & \textbf{4.98} & \textbf{11.68} & \textbf{2.33$\times$}
 & \textbf{2.27} & \textbf{2.09} \\
\bottomrule
\end{tabular}}
\vspace{-2mm}
\caption{
\textbf{Human-written and our generated rubrics by task.}
Mean numbers of criteria and hard criteria per input on the same 210 inputs. Ratio denotes the induced-to-human criterion count. Our generated rubrics contain $2.33\times$ more criteria overall and at least as many criteria on every input, while using a comparable number of hard criteria.
}
\vspace{-2mm}
\label{tab:rubric-stats}
\end{table}

Table~\ref{tab:human-correlation}  shows that VideoArgus-Full achieves higher within-instance Spearman and Kendall ranking correlations than the corresponding benchmark-specific evaluator on all five tasks. The gains are largest on T2V and TI2V, where the corresponding benchmark-specific evaluators show weak agreement with human preferences. Overall, the results show that sample-specific, criterion-level evaluation more closely recovers human model rankings across all five tasks.

We further assess statistical reliability with bootstrap confidence intervals and paired tests, reported in the appendix. On the same 210 inputs, our generated rubrics contain $2.33\times$ more criteria than human-written rubrics (11.68 vs.\ 4.98), while using a comparable number of hard criteria (2.09 vs.\ 2.27; Table~\ref{tab:rubric-stats}), indicating broader coverage without uniformly stricter requirements.

\subsection{Tool Ablation}
\begin{table}[t]
\centering
\small
\setlength{\tabcolsep}{4.0pt}
\renewcommand{\arraystretch}{1.08}

\resizebox{\columnwidth}{!}{%
\begin{tabular}{@{}l|cccc@{}}
\toprule
\textbf{Evaluation VLM}
& \textbf{A~$\leftrightarrow$~B}
& \textbf{A~$\leftrightarrow$~C}
& \textbf{B~$\leftrightarrow$~C}
& \textbf{Average} \\
\midrule
Qwen3.6-27B~\cite{qwen3.6-27b}
& 0.966 & 0.931 & 0.897 & 0.931 \\
Gemma4-31B~\cite{gemmateam2026gemma4}
& 0.966 & 0.840 & 0.851 & 0.886 \\
\midrule
\textbf{Average}
& 0.966 & 0.886 & 0.874 & \textbf{0.909} \\
\bottomrule
\end{tabular}%
}
\vspace{-2mm}
\caption{
Consistency across rubric generators, measured by task-macro rank similarity. Each entry averages the within-task Spearman correlations over five tasks, with six models ranked per task. The evaluation VLM is fixed while different rubric generators are compared. A, B, and C denote Claude Opus 4.8~\cite{anthropic2026claudeopus48}, GPT 5.6 Sol~\cite{openai2026gpt56sol}, and Gemini 3.1 Pro~\cite{googledeepmind2026gemini31pro}, respectively.
}
\vspace{-2mm}
\label{tab:rubric_generator_consistency}
\end{table}

To isolate the contribution of specialized visual tools, we replace each specialized-tool invocation in VideoArgus-Full with criterion-specific \texttt{vlm\_qa}, while keeping the rubrics, sampled frames, evaluation-VLM backbone, final-judgment prompt, and aggregation procedure fixed. As shown in Table~\ref{tab:human-correlation}, VideoArgus-Full achieves higher within-instance Spearman correlation on all five tasks and higher Kendall correlation on four tasks, although the pooled results are mixed. These results show that specialized tools provide complementary criterion-specific evidence and generally improve agreement with human within-instance rankings.

\subsection{Backbone Analysis}
\begin{table}[t]
\centering
\small
\setlength{\tabcolsep}{5.0pt}
\renewcommand{\arraystretch}{1.08}

\resizebox{0.75\columnwidth}{!}{%
\begin{tabular}{@{}l|c@{}}
\toprule
\textbf{Rubric Generator}
& \textbf{Qwen~$\leftrightarrow$~Gemma} \\
\midrule
Claude Opus 4.8~\cite{anthropic2026claudeopus48}   & 0.886 \\
GPT 5.6 Sol~\cite{openai2026gpt56sol}    & 0.954 \\
Gemini 3.1 Pro~\cite{googledeepmind2026gemini31pro} & 0.954 \\
\midrule
\textbf{Average}
& \textbf{0.931} \\
\bottomrule
\end{tabular}%
}
\vspace{-2mm}
\caption{
Consistency across evaluation VLMs, measured by task-macro rank similarity. Each entry averages the within-task Spearman correlations over five tasks, with six models ranked per task. The generated rubrics are fixed while Qwen3.6-27B~\cite{qwen3.6-27b} and Gemma4-31B~\cite{gemmateam2026gemma4} are compared as evaluation VLMs.
}
\vspace{-4mm}
\label{tab:evaluation_vlm_consistency}
\end{table}
Table~\ref{tab:rubric_generator_consistency} and~\ref{tab:evaluation_vlm_consistency} examine whether the model rankings produced by VideoArgus depend on the rubric-generation or evaluation-VLM backbone. For each pair of configurations, we compute the Spearman correlation between the rankings of six models within each task and then average equally across the five tasks. Each task is treated as an independent six-model ranking problem; raw scores and model identities are not pooled across tasks. When the evaluation VLM is fixed, changing the rubric generator yields an average rank similarity of 0.909. When the generated rubrics are fixed, replacing the evaluation VLM yields an average similarity of 0.931. In the latter comparison, each evaluation VLM is used for both criterion-specific \texttt{vlm\_qa} and final criterion judgment. The high correlations indicate that model rankings remain largely consistent across the tested backbone choices.

\subsection{Cost Analysis}

VideoArgus incurs a one-time rubric-generation cost and a recurring local evaluation cost. For each input, a rubric is generated once through an API and reused across all models and re-evaluations; subsequent scoring runs entirely on the local Qwen3.6-27B model and CV tools.

\begin{table}[h]
\centering
\setlength{\tabcolsep}{1.6mm}
\renewcommand{\arraystretch}{1.15}

\resizebox{\columnwidth}{!}{%
\begin{tabular}{l|c|cc|c}
\toprule
\multirow{2}{*}{\textbf{Task}}
& \multirow{2}{*}{\textbf{\#Input}}
& \multicolumn{2}{c|}{\textbf{Rubric-gen. tokens / input (mean)}}
& \multirow{2}{*}{\textbf{\$ / case}} \\
& & \textbf{Input} & \textbf{Output} & \\
\midrule
T2V   & 202 & 9{,}397  & 6{,}365 & 0.206 \\
TI2V  & 203 & 9{,}944  & 6{,}795 & 0.220 \\
TS2V  & 205 & 10{,}052 & 6{,}710 & 0.218 \\
TV2V  & 208 & 13{,}015 & 7{,}201 & 0.245 \\
TSV2V & 208 & 13{,}606 & 7{,}935 & 0.266 \\
\midrule
\textbf{All}
& 1{,}026
& \textbf{11{,}223}
& \textbf{7{,}007}
& \textbf{0.231} \\
\bottomrule
\end{tabular}%
}
\vspace{-2mm}
\caption{%
\textbf{Rubric-generation token cost, per task.} Mean input and output tokens of the one-time our generated rubric-generation call, averaged over all inputs of each task (the deployed rubrics), and the resulting dollar cost per case at the Opus~4.8 list price (\$5/\$25 per million input/output tokens).
}
\vspace{-4mm}
\label{tab:cost-rubric}
\end{table}

\paragraph{Rubric generation.}
As shown in Table~\ref{tab:cost-rubric}, the deployed Claude Opus~4.8 generator uses an average of 11{,}223 input and 7{,}007 output tokens per case. At \$5/\$25 per million input/output tokens, this corresponds to 0.206--0.266 per case across tasks, with an overall average of \textbf{\$0.231} over 1{,}026 inputs. Since rubrics are generated only once, this cost is amortized over all subsequent evaluations.
Using the same inputs, the average one-time cost is \textbf{\$0.323} for GPT 5.6 Sol and \textbf{\$0.125} for Gemini 3.1 Pro at their respective list prices. The difference mainly reflects output-token pricing, as rubric generation is output-dominated. VideoArgus is rubric-generator-agnostic, allowing lower-cost generators to be used with essentially no loss in alignment.

\paragraph{Evaluation.}
Scoring one generated video against its rubric requires on average 0.033 Nvidia H100 GPU hours, excluding model-loading time. This stage uses only local compute and incurs no per-result API charge, unlike several official metrics that invoke paid models such as GPT-4o or Gemini-2.5-Pro for every evaluated result.

\section{Conclusion}
We presented VideoArgus, a unified rubric-grounded framework for five video generation and editing settings. For each input instance, VideoArgus generates one output-blind, sample-specific rubric and reuses it across candidate videos. Criterion-specific evidence plans combine VLM reasoning with specialized visual tools to produce fine-grained scores, rationales, and diagnostic reports. We further introduced VideoArgus-Bench, comprising 1,026 input instances with pre-generated rubrics, and evaluated 55 model--task pairs under a shared protocol. On a separate 1,260-video human-alignment set, VideoArgus achieves stronger within-input ranking agreement than the corresponding benchmark-specific evaluators across all five tasks. Tool and backbone analyses further show that specialized visual tools provide complementary evidence and that model rankings remain largely consistent across different rubric-generation and evaluation-VLM backbones. Beyond the released benchmark, the same rubric-generation procedure can support evaluation of new user-provided input instances.

{
    \small
    \bibliographystyle{ieeenat_fullname}
    \bibliography{main}
}

\end{document}